# A Rule Based Solution to Co-reference Resolution in Clinical Text

**Ping Chen**      **David Hinote**

**Department of Computer and Mathematical Sciences**

**University of Houston - Downtown**

**Guoqing Chen**

**Baylor College of Medicine**

**VA HSR&D Center of Excellence (152)**

**2002 Holcombe Blvd. Houston TX 77030**

## ABSTRACT

**Objective:** The aim of this study was to build an effective co-reference resolution system tailored for the biomedical domain.

**Materials and Methods:** Experiment materials used in this study is provided by the 2011 i2b2 Natural Language Processing Challenge. The 2011 i2b2 challenge involves co-reference resolution in medical documents. Concept mentions have been annotated in clinical texts, and the mentions that co-refer in each document are to be linked by co-reference chains. Normally, there are two ways of constructing a system to automatically discover co-referent links. One is to manually build rules for co-reference resolution, and the other category of approaches is to use machine learning systems to learn automatically from training datasets and then perform the resolution task on testing datasets.

**Results:** Experiments show the existing co-reference resolution systems are able to find some of the co-referent links, and our rule based system performs well finding the majority of the co-referent links. Our system achieved 89.6% overall performance on multiple medical datasets.

**Conclusion:** The experiment results show that manually crafted rules based on observation of training data is a valid way to accomplish high performance in this co-reference resolution task for the critical biomedical domain.

## 1. BACKGROUND AND SIGNIFICANCE

Co-reference resolution is the process of linking together concepts that refer to the same entity. The ability to have computers automatically find this type of relation in text documents is of interest to people in the field of artificial intelligence because it can lead to having systems that can summarize texts and answer questions posed about information contained within those documents [1, 2]. Automatic summaries and question answering systems could be of great value to personnel in the healthcare industry as well. Because of these possibilities, a Natural Language Processing challenge was hosted in 2011 by the i2b2 (Informatics for Integrating Biology & the Bedside) in order to advance co-reference resolution technology for the field of automatic biomedical document analysis and understanding. Annotated data has been provided by four institutions: Partners HealthCare, Beth Israel Deaconess Medical Center, The University of Pittsburgh, and the Mayo Clinic. This data includes the original texts for medical documents, a concept file for each document that describes concepts mentioned in the texts, and chain files that identify manually created co-reference chains in each of the texts as an example of how chains are to look after processing. The concept mentions to be linked are nouns or descriptive phrases in the medical texts that represent people, actions, objects, or ideas and have been given types accordingly. There were two methods adopted by the hosts of the challenge to annotate the data sets in the i2b2 shared task. The first method is the i2b2 style annotations that include 5 concept categories: people, problems, tests, treatments, and pronouns. The other method used is ODIE (Ontology Development and Information Extraction) style annotations that include 8 categories: disease or syndrome, sign or symptom, procedure, people, other, none, laboratory or test result, and anatomical site. Each type of concept mention will only co-refer with a concept mention of the same type, with the exception of pronouns that can co-refer with any type of mention [3]. This challenge has been divided in to three tracks, and two of these the UHD team is participating in. The first track is to first find markables, and then find co-reference between them. The second track is to find co-reference relations between already marked concepts in the ODIE style of annotation. The third track is to find co-reference relations between already marked concepts in the i2b2 style of annotation.

## 2. OBJECTIVE

The aim of this study was to build an effective rule-based co-reference resolution system and compare its performance with that of some publicly available co-reference systems. For the critical biomedical domain, time and effort spent in building carefully-crafted rules could be a well justified necessity to achieve the desired performance required in practice. To fully evaluate our approach we conducted a comprehensive study that examined the performance of three publicly available general purpose co-reference resolution systems.

## 3. MATERIALS AND METHODS

The data used in this challenge came in two sets, training and test data. The training data differs from the test data by having gold standard co-reference chains included with it that could be used as a guide for constructing co-reference rules, either by hand, or by machine learning algorithms. The data sets are from the four institutions named above in section 1. Each of the institutions provided over 1000 documents[1] in that concept mentions and co-reference were marked by hand. The test data consisted of 323 documents marked using the i2b2 annotation style, and 59 documents marked using the ODIE style annotations, all of that were taken from the pool of over 1000 documents provided by the four institutions. Our method is to test 3 well-developed publicly available co-reference resolutions systems that are already built, as well as an algorithm constructed by our team, on the training data provided by i2b2. The best performing system was used to create output from the test data provided by i2b2 at the end of the challenge. Because the specifications for co-reference resolution in the i2b2 challenge were well defined and the type of data provided is specific [3], we adopted a rule-based approach for our system built in this study. For this challenge, the UHD only participated in the second and third tracks. That means our method was designed only to find co-reference chains in text documents when the gold standard concepts are already given as input. An important note is the publicly available systems are responsible for discovery of their own concept markables, whereas, the UHD rule based system is given them. In order to conduct a more equal evaluation and comparison of the four systems, the three publicly available systems would need to be capable of utilizing the gold standard data as input rather than having them rely on their own markables discovery. Attempts to do just that were made, however, since each of the systems has no facet for inputting such data, and the source code for each of the systems is not available in order to create a method to do so, it was impossible for the UHD team to have the 3 systems utilize the gold standard data as the rule based system does. When processing the data sets provided by i2b2, the gold standard concept files that came with the data were used to mark the concepts in the text documents. The system was developed by examining a sample of files, 15 per data set that we felt were representative of the data as a whole, from the pool of training data and constructing linking functions or rules, based on observation. The linking functions were checked across the unused training data set to get an idea of rules that worked, and those that did not. The system consists of six components, and uses four data sources to aid in creating co-referent links. The general architecture for our system is depicted in Figure 1.

## 3.1  Data Input and Access

The first two routines in the system are made to read in the text being examined and the concepts that are to be linked from the files provided by i2b2. The document handler breaks the text into tokens using white space boundaries, with each space character indicating the end of one word and the beginning of the next. The text is then stored in a two dimensional array where the first dimension is the line number, and the second dimension is the word number. A representation of this operation is depicted in figure 2.

The document handler controls access to this matrix and gives the system a way to easily find the location of the concepts in the text, and a way to search the words surrounding the concepts for information about the concept. The concept handler reads in each concept and stores it in an array giving each concept a number based on its position in the array. Each element in the array holds the start line, start word, end line, end word, type, and the text within each concept. The concept handler gives easy access to the attributes of each concept. An example of concept storage can be found below in figure 3.

### 3.2   Main Linker

The next routine in the algorithm is the main linker, and it matches all the concepts that are not in the person category. Every concept that passes through this linker is compared to each of the other concepts of the same type in the document and links are recorded if they meet the programmed criteria. Decisions made by this linker are binary meaning they either match or do not match. At this stage, every link that is detected is kept, that means a concept can have links to many concepts within the document, rather than at most two that is a characteristic of co-reference chains. The main linker uses string matching, the UMLS [4] (Unified Medical Language System) database, and the WordNet [5] database to determine if two concepts might have the same meaning. The main linker traverses the concept list and runs each one through its set of rules, and stores detected links in a list of pairs that is organized later on in the chain builder.

**Non-Personal Pronoun Match.** The first step with each concept is to check if it is a pronoun type. If it is a pronoun type concept and the word is "which" or "that," it is linked to the concept that immediately precedes it if the two concepts have fewer than two words between the two concepts. There are other pronouns mentioned, but any rules written for them only resulted in performance loss when testing across unused training data was conducted, we were unable to build a reliable rule for any other pronoun. Example: … deep wound culture showed MRSA which is sensitive to…

**Be Phrase Match.** The next step with each concept is to check the type of the concepts that immediately precede and follow the concept. If they are of the same type, the text in between the two concepts is examined and if it contains any words that indicate it is a "be phrase," the two concepts are linked because they are probably saying "something is something." Words and phrases that are commonly found in the "be phrases" are stored in the rule database, and were added to the database manually by the UHD team based on observations of gold standard links. Example: Resolution of organism is Methicillin - resistant Staphylococcus…

**Match by Meaning.** After the "be phrase" match, the concepts are examined and linked by their meanings. First, the concepts are conditioned by filtering out what we refer to as "common words." These common words include conjunctions (and, or, as, but, etc.), adjectives (large, blue, painful, etc.), and pronouns (he, she, it, etc.). The conjunctions and pronouns that are filtered out are chosen to be eliminated from the concept if they appear in the common words table of the rule database. Each of the words that appear in the common words table was manually placed there by the UHD team. Adjectives are detected by searches in the WordNet database. After elimination of the common words, any non-letter characters, such as punctuation and hyphens, are removed. After this conditioning, the concepts are compared to every other concept of the same type on the document in three ways.

*Head and Synonym Match.* First, every leftover word in the concept is compared to every leftover word in each of the other concepts of the same type in the document by a word comparison method. This word comparison method will declare the words a match if the first 80% of the characters in the shorter word match the same number of characters in the longer, or if they are found to be WordNet synonyms. If every word in one of the concepts is matched to a word in the other concept, a link between the two is recorded. Example: abscess ⟷ abscesses

*UMLS Match.* The second comparison is through the UMLS database. Both concepts are searched for in the MRCONSO table of the UMLS database after the conditioning, and if they are found in the database and their UMLS concept numbers match, a link between the two is recorded. Example: 'renal' and 'kidney' both have C011773 for a concept number in the UMLS database

*Acronym Match.* The third type of comparison is a check for acronyms. The first letters of each word in concepts that have two or more words are taken and are compared to whole words in other concepts, and if a whole word is found that matches either all the first letters, or some of them in order, a link is recorded.  Example: Methicillin - resistant Staphylococcus aureus ⟷ MRSA

After performing these steps, a phrase like 'Recurrent soft tissue abscess in the gluteal region' will link to 'tissue abscesses' because tissue is present in both mentions, abscess matches to abscesses by way of the head match, and since there are only two words in the second mention, all the other words in the first mention are ignored.

### 3.3  People Linker

All concept types are processed though the same path in the algorithm except for the mentions of type "person" or "people".  These mentions are processed by the people linker.  As with the main linker, all decisions made by this linker are binary.

**Identifying people mentions.** When the people linker is called to examine a document, it runs through several subroutines to identify "person" type mentions as being doctors or the subject of the document.

*Medical Personnel.* The first step performs internet searches on each concept mention.  The mention being processed is sent to a search engine, and the results are scanned for certain key words to indicate if the mention is referring to a doctor or medical personnel.  Every mention that is found to be of medical personnel is stored in a list for later use.  Example:  when 'optometrist' is sent to the search engine, it returns many results like this:

Doctors of **Optometry** and their Education | American **Optometric ...**
www.aoa.org/x5879.xml
Doctors of **Optometry** and their Education. Doctors of **optometry** are the nation's largest eye care profession, serving patients in nearly 6500 communities across **...**

These results are searched for keywords such as doctor, clinic, hospital, medical, etc.. If two or more of those types of medical keywords are present, the mention is marked as being medical personel.

*The Subject (Patient).* The second step is to find a name in the document to represent the subject of the document. The function checks each concept and if it meets these criteria:

It is not a pronoun.

It is not found to be a doctor according to the previous check.

It does not have the doctor salutation, Dr.

It has no medical title at the end, M.D.

It does not contain common words stored in the rule database such as "patient" or words that would indicate it is a family member,

That concept is marked as the subject of the document. If no such concepts that fit that criteria are found, the first occurrence of a concept that says "patient" or "pt" is marked as the subject since the patient has been the subject of the document in every document observed by the UHD team. After finding an appropriate representation of the subject, every concept that has the words "patient" or "pt" in them and no words that refer to a family member are linked to the subject concept.

*The Subject's Gender.* The third step is to find the gender of the subject, and this function simply counts the number of masculine and feminine pronouns in the document and the type that is more frequent is declared to be the gender of the subject.

**Matching People Mentions.** After gathering information about the "person" and "people" type concept mentions, the algorithm move on to actually create links between these mentions.

*Introduction Match.* If two concepts are found to be no more than 2 words apart with one starting with a doctor salutation, or ending with a medical title, and the other was marked as referring to a doctor by the internet searches or by the database which stores words that identify mentions as medical personnel

(e.g Attending), the two concepts are linked as this likely indicates an introduction of someone. Example: Please follow-up with your Optometrist, Dr. Smith 2019-01-16 at 8:30 AM.

*Partial Match.* After linking the introductions, a matching function is run that works the same way as the head matching function in the main linker. Certain words are removed from concepts, such as salutations, pronouns, titles, single letters, as well as punctuation, then, they are compared to each other. If all of the words, up to 80% of the length of the word, in each concept appear in the other concept, a link between them is recorded. This match will link people's names together, including those that appear with an initial for the first name in one instance and the full name in another. Example: You will see Edward L, Smith … on your visit to Dr. Smith's clinic

**Pronoun Linking.** The next step in the people linker is to match third person pronouns to the names to that refer to them. This is done by searching the sentence that contains pronoun concepts.

*Third Person No Proper Names in the Sentence.* If the sentence has only pronoun mentions in it, each of the pronouns in that sentence are linked to the subject concept if they are of the same gender as the subject. If it is not the same gender as the subject, the closest preceding concept that is not a pronoun is linked to it.

*Third Person With Proper Names in the Sentence.* If there is one name in the sentence, and the name's position in the sentence is before the pronoun, then it is linked to that name. If there are multiple names in the sentence, any pronoun that is the gender of the subject is linked to the subject and the others are linked to the first name in the sentence that is found to be a doctor.

*Other pronouns including First and Second Person Pronouns.* After this, any person concepts that are first person pronouns are linked together, and any second person pronouns are linked to the subject. The last step is to link any pronoun type mention that is the word "this" to the next person mention if it is within 3 words of it and the next mention is not a doctor, then, any pronoun type mention that is the word "who" is linked to the previous person type mention that is not any type of pronoun.

### 3.4 Link Filtering

After the semantic links are made in the main linker, they are passed over to filters to eliminate links that actually refer to two different entities based on clues found in the sentences surrounding the mentions in question. These clues include descriptive phrases such as dates, locations, or descriptive modifiers not included in the span of the mention. These clues are found by using regular expressions for dates and key words stored in the rule database for locations and descriptive modifiers compared by string matching. These clues are only searched for if the word preceding or following each mention is one of the key words stored in the rule database. Examples include in, on, are, is, etc. The filter portion of the algorithm also eliminates links using WordNet, any mention that is found to be an adjective with no noun included has any links to it removed.

### 3.5 Building the Chains

Once the linkers and the filter have finished their jobs, the final output is created from the "web" of links that has been made. The first concept with links is found and each link is traversed to the next concept, and each of those links is followed in a recursive fashion. A list of each concept visited is kept, and though concepts can be linked more than one time, they are added only once to the list. After every link has been examined in the "web," the list of concepts is sorted according to each concepts position in the text. Concepts that appear in the beginning of the text are at the top of the list. Once a chain is constructed, it is written to an output file in the I2B2 format.

## 4. RESULTS AND DISCUSSION

There are a number of systems publicly available for co-reference resolution. For the purpose of comparison, we conducted experiments with three widely adopted systems: BART, the Stanford co-reference system, and LingPipe, and provide their performance based on i2b2 testing data. Each system was evaluated in two ways. The first method was to compare each link with the provided co-reference chain annotations, and count it as correct only if it matches exactly with the provided annotation. With this method, single unlinked concept mentions which are not co-referent to any other mentions, called "singletons" are not considered, and links that fall in the same chain but skip an antecedent are considered incorrect. An example of this can be seen below in Figure 4.

This scoring method is referred to as "exact match" scoring and is a method we devised before receiving the i2b2 evaluation script. This method was used because the I2B2 evaluation script was not made available until late in the course of the challenge, and after use it seemed to give a better representation of the performance of the systems as we could measure individual concept type performance. The second method of evaluation is with a script provided by i2b2 that conducts 4 types of examinations of the chain output for each system: B-Cubed [6], MUC [7], Blanc [8], and CEAF [9]. Since many of concept mentions, approximately 30 to 60 percent depending on the data set, are singletons, the i2b2 evaluation script will produce a much higher score than the exact match method because it considers singletons to be correct co-reference chains. With the i2b2 evaluation script a system which produces no co-reference chains will still get all of the singletons counted as correct and will have a non-zero score that will often be much higher than the exact match score because, as previously mentioned, the exact match method does not consider singletons when scoring. Overall performance results using both methods are listed in this paper at the end of this section. The results are in the form of an f1 score, and that is the harmonic mean of precision and recall.

### 4.1 BART

Beautiful Anaphora Resolution Toolkit (BART) was developed from a project done at the 2007 Johns Hopkins Summer workshop (http://www.bart-coref.org/) [10]. Once set up, text is sent to it through a web service, and output is returned in XML format. The output contains detected concept mentions and if they belong to a chain, the chain identifier is included in the XML tag of the concept mention. A translator was created to compare the BART output to the chain files included with the input texts. Only concept mentions detected by the BART system and listed by the i2b2 annotations were considered for testing, and all other mentions and co-referent links were discarded. Individual concept type linking scores using the exact match scoring are listed in Table 1. The i2b2 evaluation script results for each training data set are shown in Table 2.

### 4.2 Stanford co-reference system

The Stanford co-reference system is an ongoing project by the Stanford Natural Processing Language Group (http://nlp.stanford.edu/software/dcoref.shtml) [11]. It uses a "Multi-pass sieve" to perform co-reference resolution, and that is a layered approach to detecting links between mentions. It starts with the strongest match first then uses more and more relaxed criteria for matches as it runs down the layers of co-referring rules. Input was supplying the raw text in a string, and output from this system comes in the form of a map stored in an array. Each element of the array holds the location, in the form of line number and word number in the text, of a source mention, and a destination mention. A simple mapping function was constructed to convert the Stanford concept locations to i2b2 concept locations. Only concept mentions that were found by the Stanford system and listed by the i2b2 annotations were considered, all other mentions and co-referent links were discarded. Individual concept type linking scores using the exact match scoring are listed in Table 1. The i2b2 evaluation script results for each training data set are shown in Table 2.

### 4.3 LingPipe

LingPipe is a suite of natural language processing tools provided by the Alias-i company as a commercial Natural Language Processing product (http://alias-i.com/lingpipe). LingPipe performs Co-reference resolution through a set of heuristic algorithms that link together mentions found by internal functions [12]. Input for the system was through command line functions specifying the location of the input text documents, and output was a text document containing xml tags surrounding discovered concept mentions and a chain identifier if the mention was found to be co-referent. A translator similar to the one used to map the BART system output was constructed to make the data useable in this study. Individual concept type linking scores using the exact match scoring are listed below in Table 1. The i2b2 evaluation script results for each training data set are shown in Table 2.

### 4.4 Our system

The reasoning behind choosing a rule based approach for the UHD algorithm was strictly because of the specific nature of the challenge. Machine learning algorithms can perform the same if not better than rule based algorithms, however, they can take much more time to construct. Rule based algorithms rely on human knowledge for their performance rather than gathering their own information. For that reason they can be quicker to build, but are less adaptable to changes in the structure and types of data. Our system could conceivably be used for other types of English texts if given concept markables in the same style as the i2b2 data. It is not restricted to only the types given for the challenge and will attempt to process concepts of any type given to it. The algorithm only looks for matching concept types before testing co-reference between the concepts. The question as to whether this algorithm would work well with a type of document other than medical documents is as of now untested. This algorithm did do well as far as adapting to new data sources in context of this competition. The University of Pittsburgh training data was released near the close of the challenge, and the UHD algorithm had shown similar performance, about 1% higher f1 score, on that data than the data on that it was being used to construct it. Individual concept type linking scores using the exact match scoring are listed in Table 1. The i2b2 evaluation script results for each training data set are shown in Table 2.

### 4.5 Combining Results

Once result data was collected, combinations of link results from the rule based system and the BART system were examined since the BART system showed the highest amount of correct link predictions. After combining the results from the two systems as a union of the sets, the statistics showed an increase of about 1% in recall but a decline of about 15% in precision, bringing the f1 score down overall. The combination of our system and BART was the only one attempted as it was felt no better gain would be achieved from the other systems in union with ours since BART performed the best, and the time it takes to test the combinations would be better spent improving our own system. Since recall of the combination of the UHD and BART systems is only 1% higher, it can be said that the UHD system found nearly all of the correct co-referent links that the other systems found with a much higher precision.

**4.6 Challenge Participation**

In order to participate in the challenge, each team participating was given test data that did not include the gold standard co-reference chains. After processing the data, each team submitted the data for evaluation by the hosts of the challenge. The system used for our submission to the challenge was the rule based system constructed by the UHD team since it showed the highest performance on the unused training data.

**Table 1:** Exact match f1 scores for the four systems on individual concept mention types in the Beth Israel, Partners Healthcare, and Mayo Clinic across unused training data.

| System | Data Set | People | Problems | Test | Treatments | All Others |
|---|---|---|---|---|---|---|
| UHD | Beth Israel | .958 | .690 | .389 | .597 | N/A |
|  | Partners Healthcare | .953 | .696 | .462 | .624 | N/A |
|  | Mayo Clinic | .593 | .667 | N/A | .500 | .453 |
| BART | Beth Israel | .590 | .202 | .166 | .300 | N/A |
|  | Partners Healthcare | .475 | .206 | .253 | .263 | N/A |
|  | Mayo Clinic | .410 | .000 | N/A | .000 | .000 |
| Stanford | Beth Israel | .205 | .076 | .000 | .096 | N/A |
|  | Partners Healthcare | .251 | .073 | .074 | .061 | N/A |
|  | Mayo Clinic | .069 | .000 | N/A | .000 | .000 |
| LingPipe | Beth Israel | .243 | .015 | .029 | .092 | N/A |
|  | Partners Healthcare | .139 | .067 | .088 | .066 | N/A |
|  | Mayo Clinic | .071 | .000 | N/A | .000 | .000 |

Table 2. i2b2 evaluation script overall f1 score results of the unused training data for all 4 systems.

| System | Beth Israel | Partners Healthcare | Mayo Clinic |
|---|---|---|---|
| UHD | .891 | .912 | .789 |
| BART | .775 | .712 | .436 |
| Stanford | .627 | .633 | .423 |
| LingPipe | .628 | .601 | .423 |

## 5. CHALLENGE RESULTS

The system the UHD team constructed had an f1 score average of .895 on all of the data sets provided for the testing. This score was the only score provided by the hosts of the competition and represents the performance of the UHD system on all of the data sets during the competition evaluation. According to the hosts of the competition, our team ranked fourth in the challenge with the top four performing systems being in a close tie for the first place. The highest performing system had an f1 score of .915.

## 6. CONCLUSION

Since the goal of the 2011 i2b2 Natural Language Processing task was to mark concept mentions as co-reference or not, the rule based system developed for this study was used to mark links in the test data released by the organization for the challenge. This decision was made based on the results from cross-checking the performance of each system on the training data provided. The results show the BART system performed the best out of the three publicly available co-reference systems tested in this study on this specific collection of data. The results also show that manually creating rules for co-reference based on observation of training data is a valid way to accomplish this co-reference task, particularly with the person type concepts in the i2b2 style annotations, and in this case performed well using the guidelines laid out by the hosts of the competition. The results listed in this paper show that the rule based system outperformed the three publicly available systems, this is due to the fact that the publicly available systems are general purpose systems designed to detect co-reference of people and named entities and the UHD rule based system was designed specifically for this challenge and these markables, and the publicly available systems must discover their own markables. The public systems should be given credit though for being able to detect co-referent links in this environment, and because they are responsible for discovering their own markables. It is not a stretch to imagine that these systems took a fair amount of time to develop, and can perform in many situations, whereas the UHD rule based system will operate in only the context of i2b2 or ODIE marked documents, which represent a variety of clinical reports from different institutions. Development cost can be higher on machine learning algorithms, like BART and the Stanford systems are. However, in specific contexts such as this competition, a high amount of performance can be achieved with the lower cost rule based algorithms. The UHD rule based algorithm could be used, theoretically, in any domain as long concepts are annotated on one of the two styles used in this challenge.

## AVAILABILITY

The UHD Co-reference Resolution system is available at

http://cms.uhd.edu/faculty/chenp/class/4319/project/.

**Funding statement**

This work was supported by National Science Foundation grant number CNS 0851984 and Department of Homeland Security grant number 2009-ST-061-C10001.

**Competing Interests Statement**

There are no competing interests from the authors.

**Contributorship Statement**

All three authors contributed significantly in the paper. Ping Chen and David Hinote designed and implemented the resolution system. Guoqing Chen worked on experiment design and result analysis.

**Figure legend**

**Figure 1.** Co-reference resolution system architecture

**Figure 2.** Representation of document handler functionality

**Figure 3.** Representation of concept handler functionality

**Figure 4.** Example of exact match scoring. Highlighted sections are concept mentions. Blue arrows are correct links, red are counted as incorrect even though it is co-reference because it skips a mention.